%% file: main.tex
\documentclass{article}
\usepackage{spconf,amsmath,graphicx}

\usepackage{graphicx}
\usepackage{amsfonts}
\usepackage{amsmath}
\usepackage{booktabs}
\usepackage{tabularx}
\usepackage{multirow}
\usepackage{color}
\usepackage{amsmath,amssymb}
\usepackage{mathrsfs}
\usepackage{marvosym}

\title{BLIND INPAINTING WITH OBJECT-AWARE DISCRIMINATION FOR ARTIFICIAL MARKER REMOVAL}
\name{\normalsize{Xuechen Guo$^1$
Wenhao Hu$^1$
Chiming Ni$^1$
Wenhao Chai$^2$
Shiyan Li$^3$\textsuperscript{(\Letter)}
Gaoang Wang$^1$\textsuperscript{(\Letter)}}}

\address{$^1$ZJU-UIUC Institute, Zhejiang University, China\\
$^2$Department of Electrical \& Computer Engineering, University of Washington, USA\\
$^3$Sir Run Run Shaw Hospital\thanks{This work is supported by the National Natural Science Foundation of China No.62106219. \Letter  means the corresponding author.}, Zhejiang University, China}

\begin{document}

\maketitle

\input{tex/abs.tex}
\input{tex/intro.tex}
\input{tex/method.tex}
\input{tex/exp.tex}
\input{tex/conclusion.tex}

% \vfill\pagebreak
\clearpage
% \section{REFERENCES}
% \label{sec:refs}

\bibliographystyle{IEEEbib}
\bibliography{ref}

\end{document}

%% file: tex/abs.tex
\begin{abstract}
Medical images often incorporate doctor-added markers that can hinder AI-based diagnosis. This issue highlights the need of inpainting techniques to restore the corrupted visual contents. However, existing methods require manual mask annotation as input, limiting the application scenarios. In this paper, we propose a novel \textbf{blind inpainting} method that automatically reconstructs visual contents within the corrupted regions without mask input as guidance. Our model includes a blind reconstruction network and an object-aware discriminator for adversarial training. The reconstruction network contains two branches that predict corrupted regions in images and simultaneously restore the missing visual contents. Leveraging the potent recognition capability of a dense object detector, the object-aware discriminator ensures markers undetectable after inpainting. Thus, the restored images closely resemble the clean ones. We evaluate our method on three datasets of various medical imaging modalities, confirming better performance over other state-of-the-art methods.

%Medical images often incorporate doctor-added markers that can hinder AI-based diagnosis. This issue highlights the need of inpainting techniques to restore the corrupted visual contents. However, existing inpainting methods require manual mask annotation as an input, limiting the application scenarios. In this paper, we propose a novel \textbf{blind inpainting} method that can automatically reconstruct visual contents within the corrupted regions without a mask input as guidance. Our model includes a blind reconstruction network and an object-aware discriminator for adversarial training. The reconstruction network consists of two branches that predict corrupted regions in an image and simultaneously restore the missing visual contents. Leveraging the potent recognition capability of a dense object detector, the object-aware discriminator ensures markers undetectable in any local regions after inpainting. As a result, the restored images can closely resemble the clean ones as much as possible. We evaluate our method on three datasets of different medical imaging modalities, and experimental results confirm it achieves better performance over other state-of-the-art methods.

\end{abstract}

%% file: tex/intro.tex
\section{Introduction}

Recent AI advancements have sparked great interest in AI-based medical diagnostics~\cite{shen2019artificial}, with medical imaging playing a crucial role~\cite{currie2019machine}. However, medical images often contain doctor-added markers that  hinder AI-based lesion detection and classification. It emphasizes the importance to restore images, especially for historical unclean data.
% , shown in Fig.~\ref{motivation} and Table~\ref{motivation}

\begin{figure}
\includegraphics[width=0.47\textwidth]{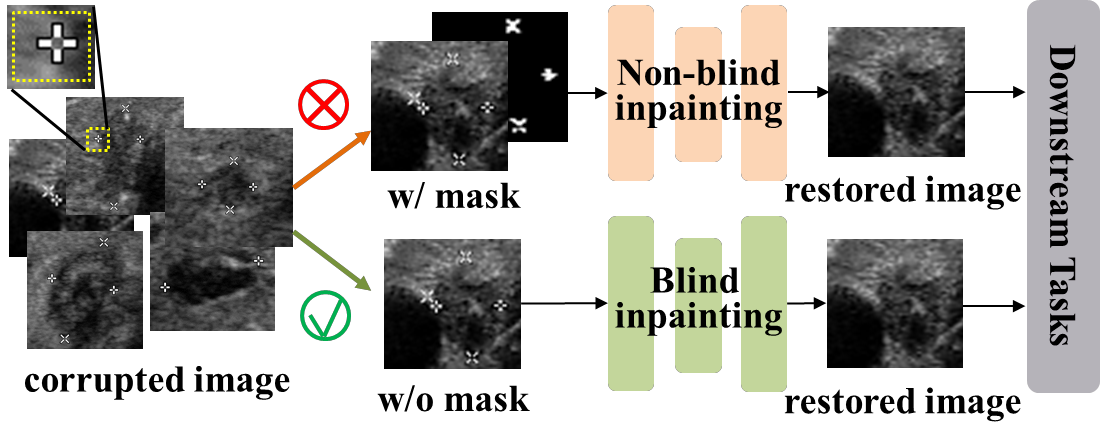}
\caption{Blind vs. Non-blind inpainting model. The blind one restores corrupted images without requiring mask annotation.}
\label{fig1}
\vspace{-5pt} 
\end{figure}

There has been substantial research into robust inpainting methods for image completion~\cite{elharrouss2020image}, including gated convolution-based~\cite{yu2019free}, transformer-based ~\cite{li2022mat}, diffusion-based ~\cite{wang2022zero} methods, \textit{etc}. 
Inpainting also finds extensive applications in medical imaging. Belli et al.~\cite{belli2018context} use adversarial training for chest X-ray image inpainting. IpA-MedGAN ~\cite{armanious2020ipa} performs well for brain MRI inpainting. Rouzrokh et al.~\cite{rouzrokh2022multitask} employ a diffusion model for brain tumor inpainting.

However, these methods often involves manual mask annotation, as shown in Fig.~\ref{fig1}, which is inconvenient, time-consuming, and error-prone. 
Blind inpainting methods~\cite{liu2019deep} offer a more practical solution, which is mask-free.
Afonso et al.~\cite{afonso2015blind} present an iterative method based on alternating minimization. 
BICNN~\cite{cai2017blind} learns an end-to-end mapping between corrupted and ground-truth pairs. VC-Net~\cite{wang2020vcnet} performs well against unseen degradation patterns with sequentially connected mask prediction and inpainting networks. However, existing works still have difficulty to localize corrupted regions, leading to sub-optimal solutions in image completion.

\indent In this work, we address the challenging blind inpainting task by creating an efficient network that is mask-free while maintaining high performance. 
Our novel framework includes a two-branch reconstruction network that predicts mask regions and implements inpainting simultaneously, and an object-aware discriminator for enhanced adversarial training. In this way, our end-to-end blind inpainting model can produce reconstructions closely resembling clean images.

\indent In summary, this paper makes the following contributions: 1) We propose a novel end-to-end blind inpainting network for artificial marker removal in medical images. 2) We design a two-branch mask-free reconstruction network for simultaneously predicting regions of markers and inpainting the corrupted visual contents. 3) We employ the object-aware discrimination by a dense object detector to ensure the restored images closely resemble clean ones. 4) Our method excels over recent blind inpainting methods on three medical image datasets of various modalities with a large margin.
%, including ultrasound (US), magnetic resonance imaging (MRI), and electron microscopy (EM), demonstrating its effectiveness. 

%% file: tex/method.tex
\begin{figure*}
    \centering
    \includegraphics[width=\textwidth]{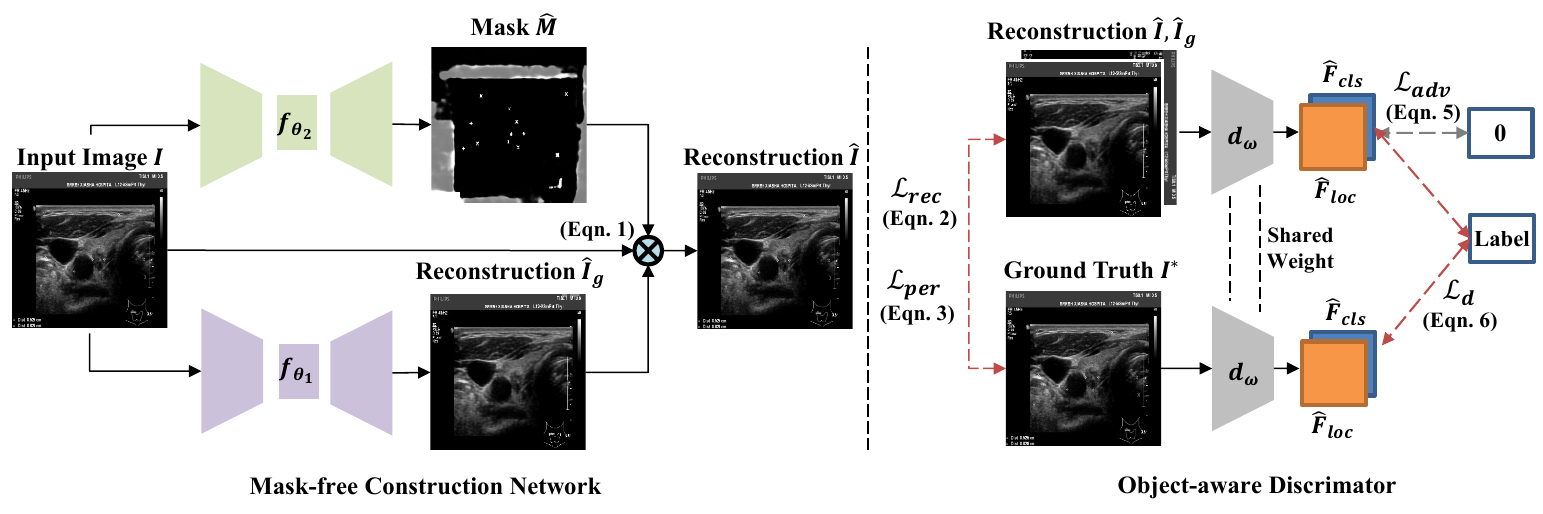}
    \caption{The proposed blind inpainting model consisted of a two-branch reconstruction network $f_{\theta}$ and an object-aware discriminator $d_{\omega}$. In $f_{\theta}$, one branch $f_{\theta_1}$ implements the inpainting task, while the other branch $f_{\theta_2}$ estimates mask of corrupted regions. $d_{\omega}$ follows the structure of dense object detectors to ensure the localization of corrupted regions.}
    \label{fig2}
\end{figure*}

\section{Method}

\subsection{Overview}

The blind image inpainting task can be described as follows. Given an input corrupted image $I$ with artificial markers, we aim to learn a reconstruction network $f_{\theta}$ to obtain a clean image $\hat{I}$ with markers removed, where $\theta$ are the network parameters to be learned. This blind inpainting task is different from the general inpainting task since the masks of corrupted regions are not provided in the inference stage. 

In the following, we minutely introduce a novel blind inpainting framework for medical imaging, as shown in Fig.~\ref{fig2}. It contains a mask-free reconstruction network and an object-aware discriminator. The reconstruction network can autonomously identify the corrupted regions and simultaneously inpaint the missing contents, eliminating the need for specific masks for target areas. In addition, the object-aware discriminator incorporates an object detector to enhance adversarial training and demonstrates the feasibility of integrating object detectors into discriminative models. 
% Details are described in following subsections.

\subsection{Mask-free Reconstruction Network}
% We employ a two-branch architecture in the reconstruction network to guide inpainting process to focus on the corrupted regions, which are unknown to the network. The two branches can localize corrupted regions and restore the missing information respectively, and thus eliminate dependency on a manual mask input.
% Specifically, the reconstruction network $f_{\theta}$ consists of two branches, $f_{\theta_1}$ for inpainting task and $f_{\theta_2}$ for mask prediction of corrupted regions. Each branch utilizes an same upsampler-convolution-downsampler structure based on gated convolution ~\cite{yu2019free}, with distinct parameters. 

We employ a two-branch architecture in the reconstruction network $f_{\theta}$ to guide the inpainting process to focus on corrupted regions, which are unknown to the network. The branch $f_{\theta_1}$ is for inpainting missing content in corrupted regions localized by the other branch $f_{\theta_2}$. This eliminats dependency on a manual mask input. Each branch utilizes an same upsampler-convolution-downsampler structure based on gated convolution ~\cite{yu2019free}, but is with distinct parameters. 
The reconstruction can be formulated as follows,

\begin{equation}
\begin{aligned}
& \hat{I}_g = f_{\theta_1}(I), \\
& \hat{M} = f_{\theta_2}(I), \\
& \hat{I} = \hat{M} \odot \hat{I}_g+(1-\hat{M}) \odot I,\\
\end{aligned}
\end{equation}
where $\odot$ represents the elementwise product.
The mask of corrupted regions is implicitly learned and the reconstruction is supervised by the clean image $I^{*}$ with the $l_1$ loss as follows,
\begin{equation}
\mathcal{L}_{\text{rec}}(\theta)=\Vert I^*-\hat{I}_g \Vert_1 + \Vert I^*-\hat{I} \Vert_1,
\end{equation}
where $\theta=\{\theta_1, \theta_2\}$.

In addition, we also constrain the feature maps of the reconstructed image with perceptual loss as follows,
\begin{equation}
\mathcal{L}_{\text{per}}(\theta)= \Vert \phi(I^*)-\phi(\hat{I}_g) \Vert_2 + \Vert \phi(I^*)-\phi(\hat{I}) \Vert_2,
\end{equation}
where $\phi(\cdot)$ is the layer activation of pre-trained VGG-16~\cite{simonyan2014very}.

\subsection{Object-aware Discrimination}
To accommodate markers of different relative sizes in corrupted images, we utilize and enhance an dense object detector such as YOLOv5~\cite{glenn_jocher_2022_6222936} to build our discriminator. This leverages the detector's powerful recognition capabilities for pixel-based classification in local regions. 
During adversarial training, the object-aware discriminator should detect artificial markers in reconstructed images as much as possible. Meanwhile, the reconstruction network should inpainting corrupted regions to blend naturally with background texture, making them less detectable as objects by the discriminator.
To enhance the discrimination in this supervision process, we define a new object category in ground-truth labels, namely ``fake marker", for marker regions in reconstructed images.

Denote the object-aware discriminator as $d_{\omega}$, where $\omega$ are the parameters to be learned. Then the output of the discriminator contains two parts, \textit{i.e.},

\begin{equation}
     \hat{F}_{\text{cls}}^{\Omega},\hat{F}_{\text{loc}}^{\Omega} =  d_{\omega}(\Omega),\quad  \Omega \in \{I^*, \hat{I}_g,\hat{I}\},
\end{equation}
where $\hat{F}_{\text{cls}}$ represents the feature maps of the classification
and $\hat{F}_{\text{loc}}$ is the localization results, including offsets and sizes.

To ensure the discriminator can be fooled, we add an adversarial loss for both $\hat{I}_g$ and $\hat{I}$, generated from the reconstruction network, \textit{i.e.},

\begin{equation}
\mathcal{L}_{\text{adv}}(\theta)= - \mathbb{E}_{\Omega \in \{\hat{I}_g,\hat{I}\}}\text{log}(1 -\hat{F}_{\text{cls}}^{\Omega})
\end{equation}
which guarantees the reconstructed image to smoothly blend with the background texture without artificial markers (objects). Set values of $\lambda_1 \sim \lambda_3$ referencing~\cite{yu2019free}.

We follow the conventional classification loss $\mathcal{L}_{\text{cls}}$ and localization loss $\mathcal{L}_{\text{loc}}$ of an anchor-based detector~\cite{glenn_jocher_2022_6222936} to train the object-aware discriminator, \textit{i.e.},
\begin{equation}
\mathcal{L}_{\text{d}}(\omega)= \sum\limits_{\Omega \in \{I^*, \hat{I}_g,\hat{I}\}} \mathcal{L}_{\text{cls}}(\hat{F}_{\text{cls}}^{\Omega};\omega)+\mathcal{L}_{\text{loc}}(\hat{F}_{\text{loc}}^{\Omega};\omega) .
\end{equation}
For the original corrupted image $I$ and the reconstructed image $\hat{I}_g$ and $\hat{I}$, the discriminator should detect the artificial markers as much as possible with the detection loss $\mathcal{L}_{\text{d}}(\omega)$. Then the total loss used for training is as follows,
\begin{equation}
\mathcal{L} = \lambda_1 \mathcal{L}_{\text{rec}}(\theta)+\lambda_2 \mathcal{L}_{\text{per}}(\theta)+\lambda_3 \mathcal{L}_{\text{adv}}(\theta)+\mathcal{L}_{\text{d}}(\omega),
\end{equation}
where $\theta$ and $\omega$ are updated iteratively.

%% file: tex/exp.tex
\section{Experiments}
\subsection{Datasets}

Our study utilizes three datasets of various medical imaging modalities. The thyroid ultrasound (\textbf{US}) dataset provided by Sir Run Run Shaw Hospital of Zhejiang University contains 414 training images, 117 validation images and 69 test images (1024×768 pixels). The images feature crosshairs and forks as doctor-added markers at lesion locations, with corresponding clean ground truth images and location labels.
The electron microscopy (\textbf{EM}) dataset sourced from the MICCAI 2015 gland segmentation challenge (GlaS)~\cite{sirinukunwattana2015stochastic} consists of 160 training images and 5 test images.  
The magnetic resonance imaging (\textbf{MRI}) dataset obtained from Prostate MR Image Segmentation Challenge~\cite{litjens2014evaluation} has 50 training images and 30 test images. To replicate the doctors' process and validate our method, we add artificial markers to EM and MRI, which initially lack them.

\subsection{Implementation Details}
We enhance the object detector YOLOv5 ~\cite{glenn_jocher_2022_6222936} to form our object-aware discriminator. And modify the generator of the non-blind inpainting model Deepfillv2 ~\cite{yu2019free} to build an improved two-branch blind reconstruction network. 
Weight factors is set as $\lambda_1=10, \lambda_2=1, \lambda_3=0.1$. Data augmentation include adding more pseudo markers randomly to input images.
To ensure a fair comparison, we maintain parameters of compared baseline models in accordance with the respective papers or codes and train until loss functions converges. Data preprocessing methods are also same.
Experiments employ a single NVIDIA RTX 3090 GPU with PyTorch. Evaluation metrics include PSNR, SSIM, and MSE. Models are optimized by Adam with learning rate $1e^{-4}$ and batch size 4.
\vspace{-10pt}
\subsection{Motivation Verification}

% \vspace{\baselineskip}
% Results show that $M_{clean}$ outperforms $M_{unclean}$.
\begin{table}[!t]
\renewcommand\tabcolsep{3pt}
\renewcommand\arraystretch{0.9}
\caption{Motivation verify: Quantitative comparison. }
\label{motivation_tab}
\centering
\begin{tabular}{l|l|cccc}
\hline
Models &
Test sets &
  \begin{tabular}[c]{@{}c@{}}P \end{tabular} &
  \begin{tabular}[c]{@{}c@{}}R\end{tabular} &
  \begin{tabular}[c]{@{}c@{}}mAP@.5\end{tabular} &
  \begin{tabular}[c]{@{}c@{}}mAP@.5:.95\end{tabular} \\ \hline
 & $V_{unclean}$   & 0.875  & 0.860 & 0.860 &  0.844\\
$M_{unclean}$ & $V_{clean}$  &0.500 &0.594 &0.556 &0.248\\
 & $V_{inpaint}$  & 0.583 &0.429 & 0.511 &0.221\\ \hline

& $V_{unclean} $  &0.780 &0.754 &0.773  &0.442\\
$M_{clean}$   & $V_{clean}$ & \textbf{0.770} & \textbf{0.696} & \textbf{0.734} & \textbf{0.425}\\
  & $V_{inpaint}$ &\textbf{0.664} &\textbf{0.719} &\textbf{0.676}  &\textbf{0.389}\\ \hline
\end{tabular}
\vspace{-6pt}
\end{table}

 % between AI diagnosis models $M_{\cdot}$ trained on unclean and clean data respectively. $M_{clean}$ outperforms $M_{unclean}$ on test set $V_{\cdot}$
\begin{figure}[!t]
    \centering
    \includegraphics[width=0.48\textwidth]{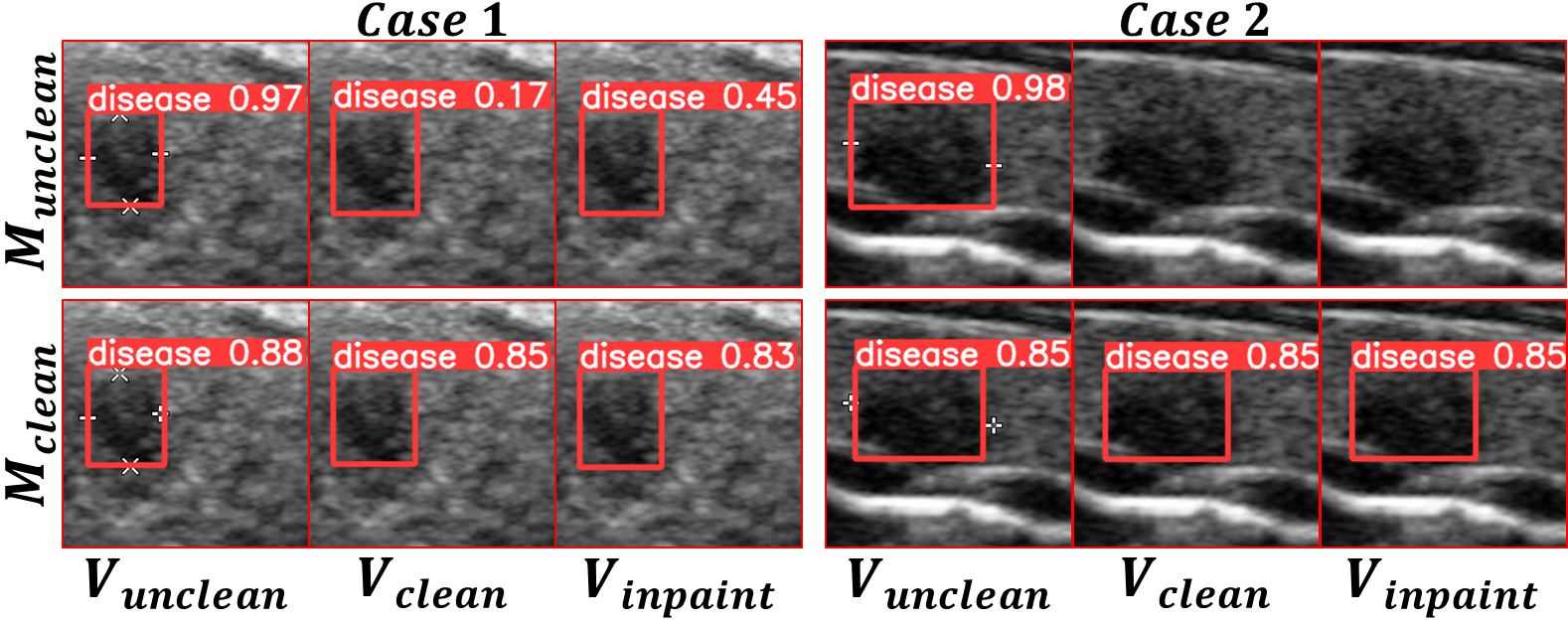}
    \caption{Motivation verify: Qualitative comparison.}
    \label{motivation}
    \vspace{-10pt} 
\end{figure}
We verify the motivation of our work by YOLOv5 for lesion detection on US dataset. First train YOLOv5 models $M_{\cdot}$ on unclean data with artificial markers and clean data respectively. Use $V_{\cdot}$ as test sets and process $V_{unclean}$ by our inpainting model to obtain $V_{inpaint}$.
Shown in Fig.~\ref{motivation} and Table~\ref{motivation_tab}, 
% excluding the case where $M_{unclean}$ performs well on $V_{unclean}$ just because conspicuous markers simplify feature learning, 
$M_{unclean}$ detects lesions relying on marker recognition, rather than understanding medical  semantics as $M_{clean}$. 
It proves the negative impact of unclean data on AI diagnostics.  

% We verify our work's motivation by YOLOv5 for lesion detection on US dataset. First train YOLOv5 models $M_{unclean}$, $ M_{clean}$ on unclean data with artificial markers and clean data respectively. Then use $V_{\cdot}$ as test sets. $V_{inpaint}$ is obtained by processing $V_{unclean}$ using our inpainting model. As shown in Fig.~\ref{motivation} and Table~\ref{motivation_tab}, $M_{unclean}$ underperforms $M_{clean}$. It proves the negative impact of unclean data on AI diagnostics when testing and extensive historical data training. 

\subsection{Main Results}

\begin{table}[!t]
\renewcommand\tabcolsep{3.1pt}
\renewcommand\arraystretch{0.9}
\caption{Quantitative comparison between our method, VCNet~\cite{wang2020vcnet}, MPRNet~\cite{zamir2021multi} and UNet~\cite{ronneberger2015u} (mean±s.d). In parentheses are metrics further calculated \textbf{only within mask areas}.}
\label{tab1}
\centering
\begin{tabular}{l|l|cccc}
\hline
Data &
Methods &
  \begin{tabular}[c]{@{}c@{}}PSNR~$\uparrow$ \end{tabular} &
  \begin{tabular}[c]{@{}c@{}}SSIM~$\uparrow$\end{tabular} &
  \begin{tabular}[c]{@{}c@{}}MSE~$\downarrow$\end{tabular} \\ \hline
 & MPRNet   & $37.877_{\pm3.289}$ &$0.995_{\pm0.002} $ & $13.027_{\pm10.201}$ \\
  &    & ($13.478$) &($0.429) $ & ($3213.933$) \\
& UNet    & $35.262_{\pm1.319}$ &$0.985_{\pm0.004} $ & $20.499_{\pm9.442}$\\
US &   & ($14.899$) &($0.419) $ & ($2280.374$)\\
 & VCNet  & $36.891_{\pm1.425}$ &$0.971_{\pm0.012}$  &$14.442_{\pm6.910}$\\
  &   & ($28.988$) &($0.801 $) & ($\textbf{87.293}$) \\
 & Ours  & $\textbf{47.673}_{\pm5.415}$ & $\textbf{0.999}_{\pm0.001}$ & $\textbf{2.633}_{\pm5.856}$  \\ 
 &   & ($\textbf{30.016}$) & ($\textbf{0.855}$) & ($103.111$)  \\ \hline
 
 & MPRNet  & $34.860_{\pm1.992}$ &$0.991_{\pm0.001} $ & $23.298_{\pm9.599}$ \\
 &   & ($17.692$) &($0.627) $ & ($1226.490$) \\
& UNet     & $29.736_{\pm2.004}$ &$0.961_{\pm0.012} $ & $75.659_{\pm29.296}$\\
MRI  &    & ($18.021$) &($0.625) $ & ($1003.576$) \\
  & VCNet     & $31.315_{\pm1.405}$ &$0.947_{\pm0.029} $ & $63.405_{\pm18.734}$\\
  &     & ($21.117$) &($0.705 $) & ($423.108$)\\
  & Ours  & $\textbf{40.049}_{\pm7.004}$ & $\textbf{0.994}_{\pm0.003}$& $\textbf{7.153}_{\pm9.627}$  \\
  &   & ($\textbf{26.159}$) & ($\textbf{0.821}$) & ($\textbf{203.967}$) \\ \hline

& MPRNet   & $35.184_{\pm1.368}$ &$0.991_{\pm0.002} $ & $20.505_{\pm6.460}$ \\
&   & ($18.354$) &($0.702 $) & ($1004.690$)\\
& UNet    & $34.239_{\pm0.847}$ &$0.984_{\pm0.001} $ & $24.881_{\pm4.931}$\\
CM   &   & ($19.472$) &($0.707) $ & ($1015.378$)\\
  & VCNet    & $32.230_{\pm0.350}$ &$0.956_{\pm0.007} $ & $39.016_{\pm3.098}$\\
  &   & ($22.268$) &($0.718 $) & ($387.710$)\\
  & Ours & $\textbf{41.419}_{\pm1.902}$& $\textbf{0.997}_{\pm0.001}$ & $\textbf{2.595}_{\pm1.284}$\\
   &  &  ($\textbf{28.437}$)& ($\textbf{0.839}$) & ($\textbf{165.442}$) \\ \hline
\end{tabular}
\end{table}

% %arxiv:
% \begin{table}[!h]
% \renewcommand\tabcolsep{7pt}
% \caption{Further quantitative comparison between ours, VCNet, MPRNet and UNet. Metrics are calculated \textbf{solely in mask areas}, represented by \textbf{rectangular boxes} derived from ground-truth labels.}
% \label{tabA}
% \centering
% \begin{tabular}{c|l|cccc}
% \toprule
% Dataset &
% Methods &
%   \begin{tabular}[c]{@{}c@{}}PSNR~$\uparrow$ \end{tabular} &
%   \begin{tabular}[c]{@{}c@{}}SSIM~$\uparrow$\end{tabular} &
%   \begin{tabular}[c]{@{}c@{}}MSE~$\downarrow$\end{tabular} \\ \midrule
%  & MPRNet   & $13.478$ &$0.429 $ & $3213.933$ \\
% US & UNet  & $14.899$ &$0.419 $ & $2280.374$ \\
%  & VCNet  & $28.988$ &$0.801 $ & $\textbf{87.293}$ \\
%  & Ours  & $\textbf{30.016}$ & $\textbf{0.855}$ & $103.111$  \\ \midrule
 
%  & MPRNet  & $17.692$ &$0.627 $ & $1226.490$ \\
% MRI  & UNet   & $18.021$ &$0.625 $ & $1003.576$ \\
%   & VCNet    & $21.117$ &$0.705 $ & $423.108$\\
%   & Ours  & $\textbf{26.159}$ & $\textbf{0.821}$ & $\textbf{203.967}$ \\ \midrule

% & MPRNet  & $18.354$ &$0.702 $ & $1004.690$\\
% CM   & UNet  & $19.472$ &$0.707 $ & $1015.378$\\
%   & VCNet  & $22.268$ &$0.718 $ & $387.710$\\
%   & Ours &  $\textbf{28.437}$ & $\textbf{0.839}$ & $\textbf{165.442}$ \\
% \bottomrule
% \end{tabular}
% \end{table}

\begin{figure}[!t]
    \centering
    \includegraphics[width=0.48\textwidth]{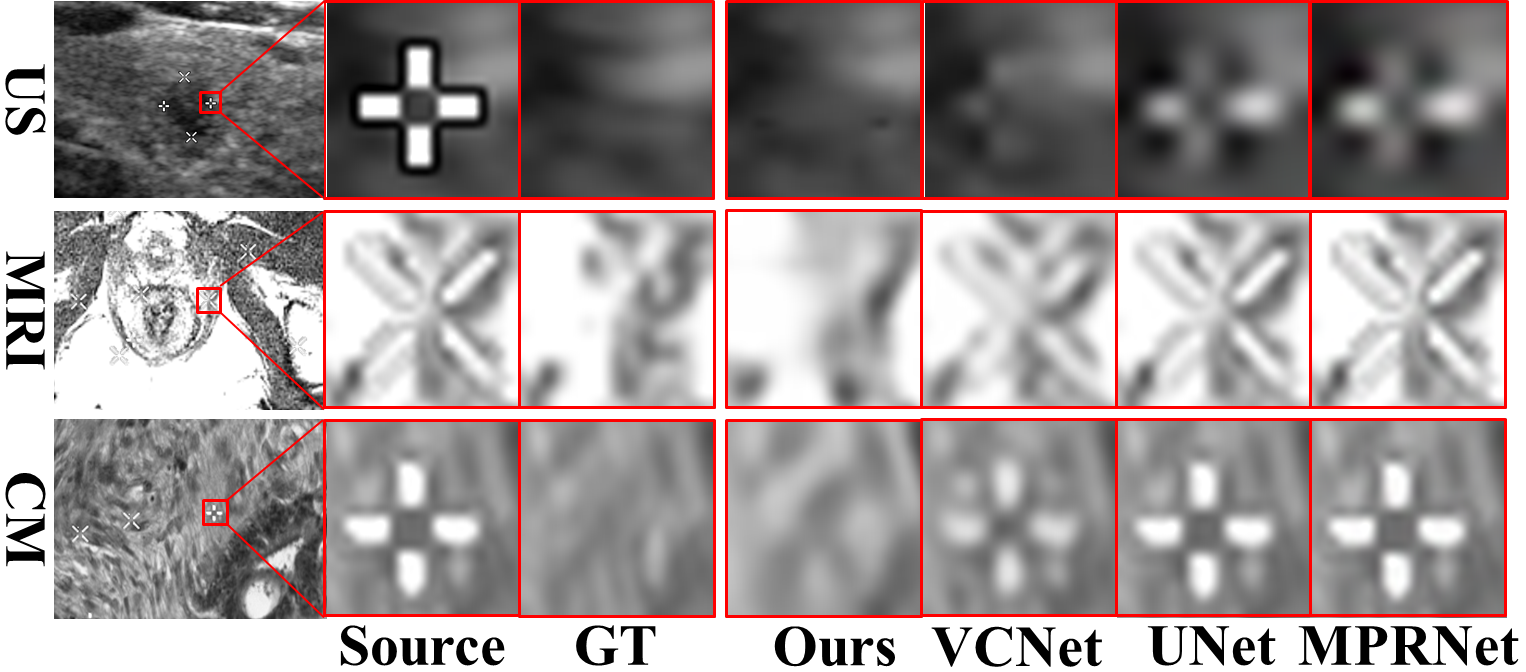}
    \caption{Qualitative comparison. Our model generates visually appealing results. Other models exhibit varying levels of restoration failure.}
    \label{fig3}
\end{figure}

\begin{figure}[!t]
\includegraphics[width=0.47\textwidth]{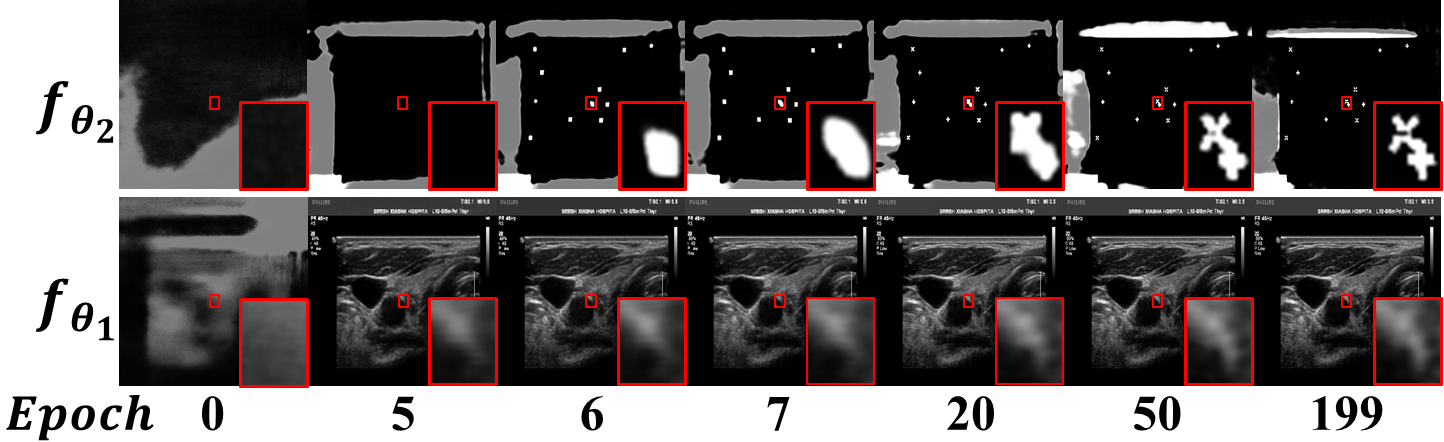}
\caption{Results of two-branch generator included mask prediction branch $f_{\theta_2}$ and inpainting branch $f_{\theta_1}$ when training.}\label{figC}
\vspace{-10pt}
\end{figure}

We evaluate our method through comparisons with recent blind inpainting framework VCNet~\cite{wang2020vcnet} and SOTA reconstruction networks MPRNet~\cite{zamir2021multi} and UNet~\cite{ronneberger2015u}. 
Table~\ref{tab1} quantitatively compares our model to baselines, demonstrating superior restoration ability with statistically significant improvements. Metrics are further calculated within mask areas determined by ground-truth location labels, confirming our method's effectiveness.
Fig.~\ref{fig3} demonstrates a qualitative superiority of our method over VCNet in terms of restoration. Additionally, results from UNet and MPRNet suggest that denoising and general reconstruction methods are inadequate for this task.
And Fig.~\ref{figC} depicts the learning process of the two-branch generator for mask prediction and inpainting.

\subsection{Ablation Study }
\begin{figure}[t]
    \centering
    \includegraphics[width=0.48\textwidth]{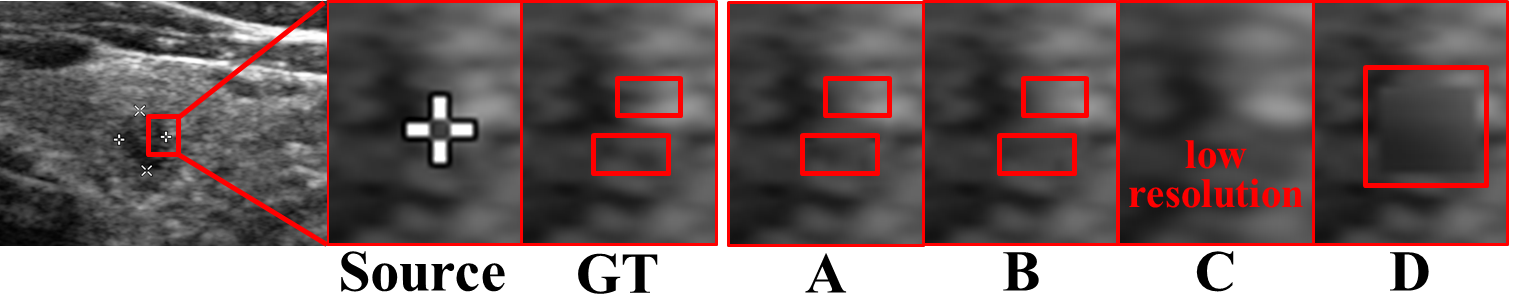}
    % \vspace{-6pt}
    \caption{Qualitative ablation study. Complete ``A" gives visually appealing results. ``B" loses fine texture details. ``C" has low-quality resolution. ``D" shows restoration degradation.}
    \label{fig4}
    \vspace{-8pt}
\end{figure}

We compared our implementation with other different structures on US dataset, as shown in Table~\ref{tab2} and Fig.~\ref{fig4}.

\begin{table}[!t]
\renewcommand\tabcolsep{8pt}
\renewcommand\arraystretch{0.883}
\caption{Ablation study on US dataset. ``A" is our complete model. ``B" replaces our object-aware discriminator with the one in Deepfillv2. ``C" replaces our two-branch reconstruction network with a single branch one. ``D" is a two-stage non-blind inpainting solution with YOLOv5 and Deepfillv2.}
\label{tab2}
\centering
\begin{tabular}{c|ccc}
\hline
Type &

  \begin{tabular}[c]{@{}c@{}}PSNR~$\uparrow$ \end{tabular} &
  \begin{tabular}[c]{@{}c@{}}SSIM~$\uparrow$\end{tabular} &
  \begin{tabular}[c]{@{}c@{}}MSE~$\downarrow$\end{tabular} \\ \hline

A & $\textbf{47.673}_{\pm5.415}$
 & $\textbf{0.999}_{\pm0.001}$ & $\textbf{2.633}_{\pm5.856}$
 \\ \hline
B &$33.283_{\pm2.023}$ & $0.984_{\pm0.006}$  &$33.948_{\pm16.306}$\\
C &$29.306_{\pm2.131}$ & $0.883_{\pm0.038}$  &$87.551_{\pm52.855}$ \\
D  & $43.551_{\pm3.014}$ &$0.998_{\pm0.001}$  &$4.583_{\pm9.094}$\\
\hline
\end{tabular}
\vspace{-6pt}
\end{table}

\vspace{5pt}
\noindent \textbf{Object-aware Discrimination.}
We replace our discriminator with the one in SN-PatchGAN from Deepfillv2 as ``B" in Table~\ref{tab2}. Performance degrades in all metrics, particularly in MSE and PSNR, suggesting loss of fidelity. Fig.~\ref{fig4} highlight our complete model’s success with robust recognition capability to identify markers after enhanced adversarial training.

\vspace{5pt}
\noindent \textbf{Two-branch Reconstruction Network Structure.}
Replace our two-branch reconstruction network with a single branch one as model ``C". Table~\ref{tab2} indicates that our complete model ``A" outperforms model ``C" with a 62.67\% improvement in PSNR. Fig.~\ref{fig4} illustrates that ``C" loses texture details, while ``A" produces visually superior results, thanks to the mask prediction branch focusing on corrupted region during fusion. 

\vspace{5pt}
\noindent \textbf{Comparison with the Two-Stage Non-blind Baseline.}
The original YOLOv5~\cite{glenn_jocher_2022_6222936} + Deepfillv2~\cite{yu2019free} two-stage non-blind inpainting network is compared as a baseline ``D".
Both quantitative and qualitative results depict an obvious degradation in texture restoration compared to our end-to-end blind inpainting model. It confirms the superiority of our approach.

%% file: tex/conclusion.tex
\section{Conclusion}
In this work, we propose a novel blind inpainting method with a mask-free reconstruction network and an object-aware discriminator for artificial marker removal in medical images. 
It eliminates dependency on the technical manual mask input for corrupted regions in an image. And we demonstrate the practicability of employing an dense object detector to the discriminator. We validate our method on multiple medical image datasets such as US, EM, and MRI, verifying its efficiency and robustness for this task. 
For future works, we plan to combine diffusion models in the reconstruction network and validate the performance in large hole blind inpainting.